\documentclass[11pt, a4paper, logo, copyright, nonumbering]{deepmind}

\pdftrailerid{redacted}

\makeatletter
\renewcommand\bibentry[1]{\nocite{#1}{\frenchspacing\@nameuse{BR@r@#1\@extra@b@citeb}}}
\makeatother

\usepackage{kantlipsum, lipsum}
\usepackage{dsfont}
\usepackage{dm-colors}

\usepackage[authoryear, sort&compress, round]{natbib} 

\graphicspath{{figures/}}

\title{Statistical Discrimination in Learning Agents}

\correspondingauthor{duenez@deepmind.com}

\reportnumber{2115} % Leave blank if n/a 

\renewcommand{\today}{2021-10-14} 

\author[1]{Edgar~A.~Du\'e\~nez-Guzm\'an}
\author[1]{Kevin~R.~McKee}
\author[1]{Yiran~Mao}
\author[1]{Ben~Coppin}
\author[1]{Silvia~Chiappa}
\author[1]{Alexander~S.~Vezhnevets}
\author[1]{Michiel~A.~Bakker}
\author[1]{Yoram~Bachrach}
\author[2]{Suzanne~Sadedin}
\author[1]{William~Isaac}
\author[1]{Karl~Tuyls}
\author[1]{Joel~Z.~Leibo}

\affil[1]{DeepMind}
\affil[2]{Independent researcher}

\begin{abstract}
  Undesired bias afflicts both human and algorithmic decision making, and may be especially prevalent when information processing trade-offs incentivize the use of heuristics. One primary example is \textit{statistical discrimination}---selecting social partners based not on their underlying attributes, but on readily perceptible characteristics that covary with their suitability for the task at hand. We present a theoretical model to examine how information processing influences statistical discrimination and test its predictions using multi-agent reinforcement learning with various agent architectures in a partner choice-based social dilemma. As predicted, statistical discrimination emerges in agent policies as a function of both the bias in the training population and of agent architecture. All agents showed substantial statistical discrimination, defaulting to using the readily available correlates instead of the outcome relevant features. We show that less discrimination emerges with agents that use recurrent neural networks, and when their training environment has less bias. However, all agent algorithms we tried still exhibited substantial bias after learning in biased training populations.
\end{abstract}

\begin{document}

\maketitle 

\section{Introduction}

Choosing good social partners is one key to successful human cooperation~\citep{henrich2017secret}, but accurately evaluating a potential partner requires acquiring, integrating and processing information across a variety of situations~\citep{brosnan2002proximate, schino2009reciprocal}. Information processing trade-offs create pressure to learn and apply cheap heuristics in partner choice~\citep{schino2010primate}. These heuristics, however, can produce undesired biases. Such biases are evident in both human behavior~\citep{greenwald1995implicit, devine1989stereotypes} and machine learning algorithms~\citep{caliskan2017semantics, zuiderveen2018discrimination}, and have clear negative effects on economic efficiency and human wellbeing~\citep{bielby1986men,schwab1986statistical,maitzen1991ethics,ayres1991fair,fang2011theories, li2020hiring}. However, the process of their development has not been widely studied.

One formalism used to understand the emergence of such heuristics is \emph{statistical discrimination}. Statistical discrimination occurs when a decision maker must assess an individual's suitability as a partner but lacks immediate access to features that are causally relevant to that choice. Instead the decision maker bases its assessment on features that are readily available and, in their experience, correlate with the ones of interest~\citep{arrow1972some, phelps1972statistical}. Statistical discrimination stands in contrast to taste-based discrimination, where decision makers have an intrinsic preference for one group over another~\citep{guryan2013taste}.

Models of statistical discrimination do not attempt to capture the full complexity of racial, gender, and other types of discrimination in human societies, which likely involve a wide range of cognitive phenomena such as \emph{in-group favoritism}~\citep{balliet2014ingroup}, \emph{group entitativity}~\citep{yzerbyt1998group}, and more. Nonetheless, it is possible that multi-generational patterns of systematic social exclusion could be initiated and/or supported by statistical discrimination~\citep{arrow1998has}. 

Traditional statistical discrimination models assume a decision maker evaluates a referent individual in a one-shot interaction where only perceptible information is available. In reality, decision makers can improve their knowledge of the outcome-relevant characteristics of social partners in a gradual manner. For instance, even in the employment setting where statistical discrimination was originally formulated by~\cite{arrow1972some}, employers in the real world refine their assessment of job applicants with successive filters and stages. Employers use r\'esum\'e screening, multiple rounds of interviews, internships and other provisional employment opportunities to improve their knowledge of the applicant's suitability for a job. In other words, real-world situations are temporally extended, require acquiring, processing and integrating information across time, and comprise a complex tapestry of uncertainty, repeated interactions, and risk management. Limits to time, information processing ability, or the reliability of signals of partner quality will therefore impact the accuracy of decision making, and produce higher bias. Yet, there currently exist no theoretical models to understand the interplay of such limits.

In this work, we introduce a new game theoretical framework for statistical discrimination which advances the state-of-the-art in three ways: (i) allows for sequential decision making; (ii) provides individuals with the ability to choose their partners; and (iii) allows for evaluation of partners based on direct or indirect experience. We then validate this model on a temporally and spatially extended multiagent reinforcement learning environment where agents with varying information processing abilities engage in repeated social interactions with partner choice.

\begin{figure*}
  \centering
  \includegraphics[scale=0.6]{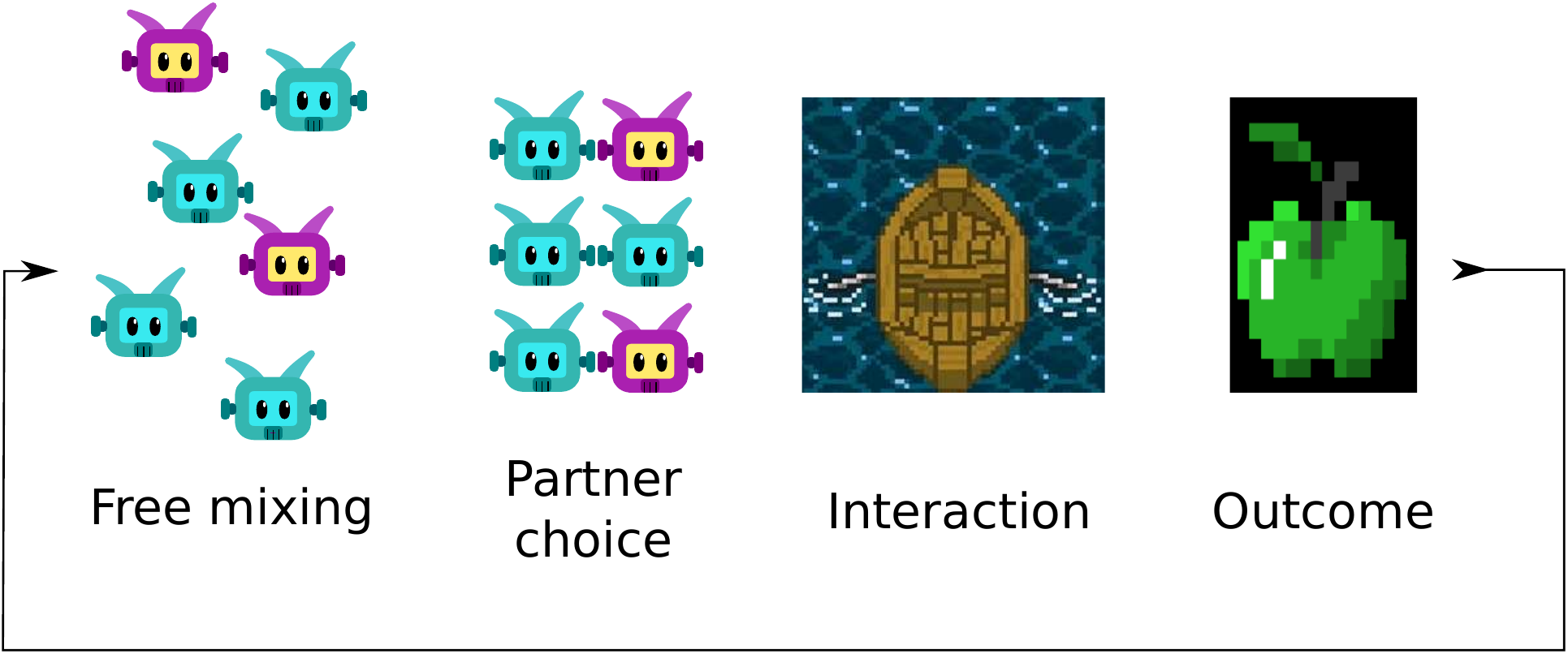}
  \caption{Cooperative iterations follow a pattern of repeated phases: During free mixing, individuals have access to observable features, but not directly their suitability as partners. Individuals form pairs based on what information is available to them and proceed to interact. The outcome of the interaction is a pair of payoff (or rewards) assigned to the individuals in a pair.\label{fig:schematic}}
\end{figure*}

\subsection{Reinforcement learning as a model of human behavior}

Recently, multiagent reinforcement learning (MARL) has emerged as a framework for the study of social interactions~\citep{tuyls2005evolutionary,kleiman2016coordinate,leibo2017multiagent,peysakhovich2018towards} by generalizing traditional models of social behaviour to more realistic scenarios in two important ways. First, it enables study of more complex environments than the ones covered by traditional game theory methods; the environments can be spatial, and temporally extended, grounding the abstract conceptions of actions and interactions in a rich world. Second, it enables the study of agents in those rich environments whose decision processes are limited by learning algorithms and information processing ability, and whose interactions are grounded by the laws of their worlds.

In contrast to many game theoretic models that assume optimal behaviour, models of bounded rationality explore the role of limited information processing ability and biases~\citep{simon1957models} in the dynamics and behaviors of a population. MARL offers a fresh approach to modeling bounded rationality, focusing on the learning dynamics, where the bounds are implicit to the learning process~\citep{leibo2017multiagent, mckee2021deep}.

MARL has been used to study a variety of social situations, including social dilemmas~\citep{leibo2017multiagent, lerer2017maintaining, hughes2018inequity}, norm and convention formation~\citep{lerer2019learning, koster2020model, 10.5555/3463952.3464123}, communication~\citep{foerster2016learning, lazaridou2020multi}, and many more. A recent article explored partner choice in an abstract iterated social dilemma environment~\citep{anastassacos2020partner}, validating that partner choice can indeed promote cooperation in learning agents.

\subsection{Partner choice in human societies}

Partner choice is a critical mechanism for cooperation in nature~\citep{noe1994biological, barclay2007partner, fu2008reputation}, and evaluation of social partners is postulated as a major driving force in human evolution~\citep{dunbar1998social}. Previous theoretical work has established that partner choice is a mechanism that can promote cooperation \citep{Santos2006, gintis2008strong}. Ostracism has been observed to promote and maintain cooperation in cultural contexts ranging from hunter gatherers and small-scale pastoral societies~\citep{soderberg2016anthropological}, to ancient Athenians~\citep{ouwerkerk2005avoiding}, as well as in experimental studies of contemporary Westerners~\citep{feinberg2014gossip}. Given its cross-cultural ubiquity, it is plausible that evolved cognitive mechanisms for partner choice contribute to systemic social exclusion~\citep{tooby1988evolution, kurzban2001evolutionary}.

Due to limits of cognition and information processing, the mechanisms involved in partner choice may go beyond the (rational but undesirable) statistical discrimination and end up being applied irrationally. This is especially likely where social evaluations are rapid; humans make more mistakes when making decisions under time pressure~\citep{rubinstein2013response}, and these results transfer to social interactions~\citep{evans2015fast} including in responses to discrimination~\citep{greenwald1995implicit, fitzgerald2017implicit}. In primates, there is evidence that reciprocity is driven by relatively simple cognitive processes~\citep{brosnan2002proximate, schino2010primate}. In humans, \emph{implicit bias} is a measurable and unconscious process that biases perception of social partners based on their perceived group membership \citep{greenwald1995implicit}. Reduction of implicit bias is possible, but appears to require awareness of the bias, and a conscious effort to diminish it~\citep{devine2012long} (cf. \citep{lai2016reducing}).

\section{A model of discrimination with partner choice}

We abstract the scenario of choosing desirable social partners to engage in potentially mutually beneficial interactions as an iterated game. Each iteration consists of several phases (see Figure~\ref{fig:schematic}). A \emph{free mixing} phase where individuals can observe some characteristics of others that might or might not correlate with actual performance. Crucially, the actual partner quality cannot be directly observed in this phase. A \emph{partner choice} phase where individuals pair up, and commit to interact to the exclusion of all other possible social partners. After choosing partners, the pairs continue to the \emph{interaction} phase, where the outcome relevant behaviors are expressed. The \emph{outcome} phase assigns payoffs (or rewards) to each of the individuals based on how they, and their partner, behaved during the interaction phase.

Iterated social dilemmas are the dominant framework to study the dynamics of social relationships~\citep{axelrod1981evolution, Nowak1992}. They consist of a population of individuals engaging in repeated pair-wise social interactions. We extend this framework by introducing partner choice \citep{stanley1993iterated, hauk2001leaving}. For simplicity, we assume that if both individuals in a pair were happy with the outcome of their previous interaction, they can and will be able to pair up again at the next partner choice phase. Individuals can choose to unilaterally end a relationship by simply choosing someone else. In addition to observing the outcome of their immediate interaction, individuals might be able to observe the outcomes of other interactions in the population. Constraining an individual's access to their outcomes, or the outcomes of other pairs, is how we model information processing ability in our analytical model. In the reinforcement learning model, this ability to perceive and process outcomes of interactions between other individuals is grounded on the observations of the environment (see Section \ref{sec:rl}).

As is typical in iterated social dilemmas, we model interactions as a(n instantaneous) matrix game. In an interaction, each individual has access to two strategies: \emph{cooperate} or \emph{defect}. Individuals then simultaneously choose one strategy, and receive a payoff depending on a payoff matrix $A$. In this work we use the payoff matrix of \emph{Stag Hunt}~\citep{rousseau2009discourse, skyrms2001stag},
\begin{equation}
\begin{array}{cc}
 & \begin{array}{cc}
C & D\end{array}\\
\begin{array}{c}
C\\
D
\end{array} & \left(\begin{array}{cc}
R & S\\
T & P
\end{array}\right)
\end{array}
\label{eq:staghunt}
\end{equation}
where $R > T \ge P > S$. An individual playing strategy $i$ interacting with one playing strategy $j$ receives payoff $A_{i,j}$. So, for instance, if both individuals cooperate, both get the payoff $R$, whereas if only the first one cooperates, they get $S$, and their partner gets $T$.

We chose \emph{Stag Hunt} instead of the \emph{Prisoner's Dilemma} as the underlying dynamics of the social interactions because our focus here is on situations where cooperation is prevalent in the population. The emergence and maintenance of cooperation has been extensively studied elsewhere~\citep{Nowak2006, Fletcher2009, Henrich2010fairness}). In contrast to the \emph{Prisoner's Dilemma} where there is always an incentive to defect, the incentive in \emph{Stag Hunt} is to coordinate with your partner. Mutual cooperation is preferred over mutual defection, and once mutual cooperation has been achieved, there is no incentive to deviate. Because of these properties, \emph{Stag Hunt} is regarded as better capturing the dynamics of long term cooperation~\citep{skyrms2004stag}.

Individuals have a set of perceptible attributes that may or may not correlate with the strategy they play during an interaction. We consider a focal individual living through $k$ social iterations. We assume a basic level of rationality, that is, they would not choose to defect or end a relationship when they are in a mutually cooperative partnership. We model their different levels of rationality based on their ability to react to the behaviour of their partners, the behaviours of other individuals in the population, and the specific partner sampling strategy they employ. In particular, we consider two broad types of behaviors: \emph{unconditional}, which either cooperates or defects, regardless of partner, and never ends relationships; and \emph{reciprocating}, which cooperates when partner cooperates, otherwise ends the relationship. We also consider two partner sampling styles: \emph{visual}, which samples exclusively based on perceptible attributes; and \emph{aware}, which samples first based on perceptible attributes, but observes all interactions in the population, thereafter, samples cooperators exclusively.

Intuitively, an unconditional behavior is less sophisticated than a reciprocating one. And reciprocating behaviors that rely on \emph{visual} sampling are less cognitively demanding than ones with \emph{aware} sampling. We also refer to an \emph{omniscient} behavior which is provided as an upper-bound, representing the maximum possible payoff. An omniscient individual knows the outcome-relevant features, and ignores any other information.

We denote by $\rho$ the probability that, on the first interaction, the focal individual chooses a partner that cooperates. This quantity depends on a wide range of factors, including the individual's partner choice policy, the population composition, and the correlation between perceptible features and behaviors. This quantity allows us to write the payoffs of individuals expressing different policies in a compact and intuitive way. For a full explanation of this quantity and how it maps to individuals' information processing abilities, please refer to the Supplementary Material.

Behaviours and sampling styles are largely orthogonal, except that an \emph{unconditional aware} individual would be hard to distinguish from a \emph{visual aware} one. Therefore, we only consider the following four policies:

\begin{enumerate}
    \item \textbf{Visual unconditional}: samples partners of only one color, and unconditionally cooperates with them ($\pi_{\text{VU}}$).
    \item \textbf{Visual reciprocator}: samples partners of only one color, cooperates first with them, but ends the relationship if they defect ($\pi_{\text{VR}}$).
    \item \textbf{Aware reciprocator}: samples first a partner of a specific color and cooperates with them, if they defect, exclusively samples cooperators from then on (the individual is aware of the behaviors of others from the second iteration) ($\pi_{\text{AR}}$).
    \item \textbf{Omniscient}: always samples cooperators and cooperates with them ($\pi_{\text{O}}$).
\end{enumerate}
which stand in as our buckets for the information processing of individuals, from low to high, respectively.

\begin{figure}
  \centering
  \includegraphics[scale=0.6]{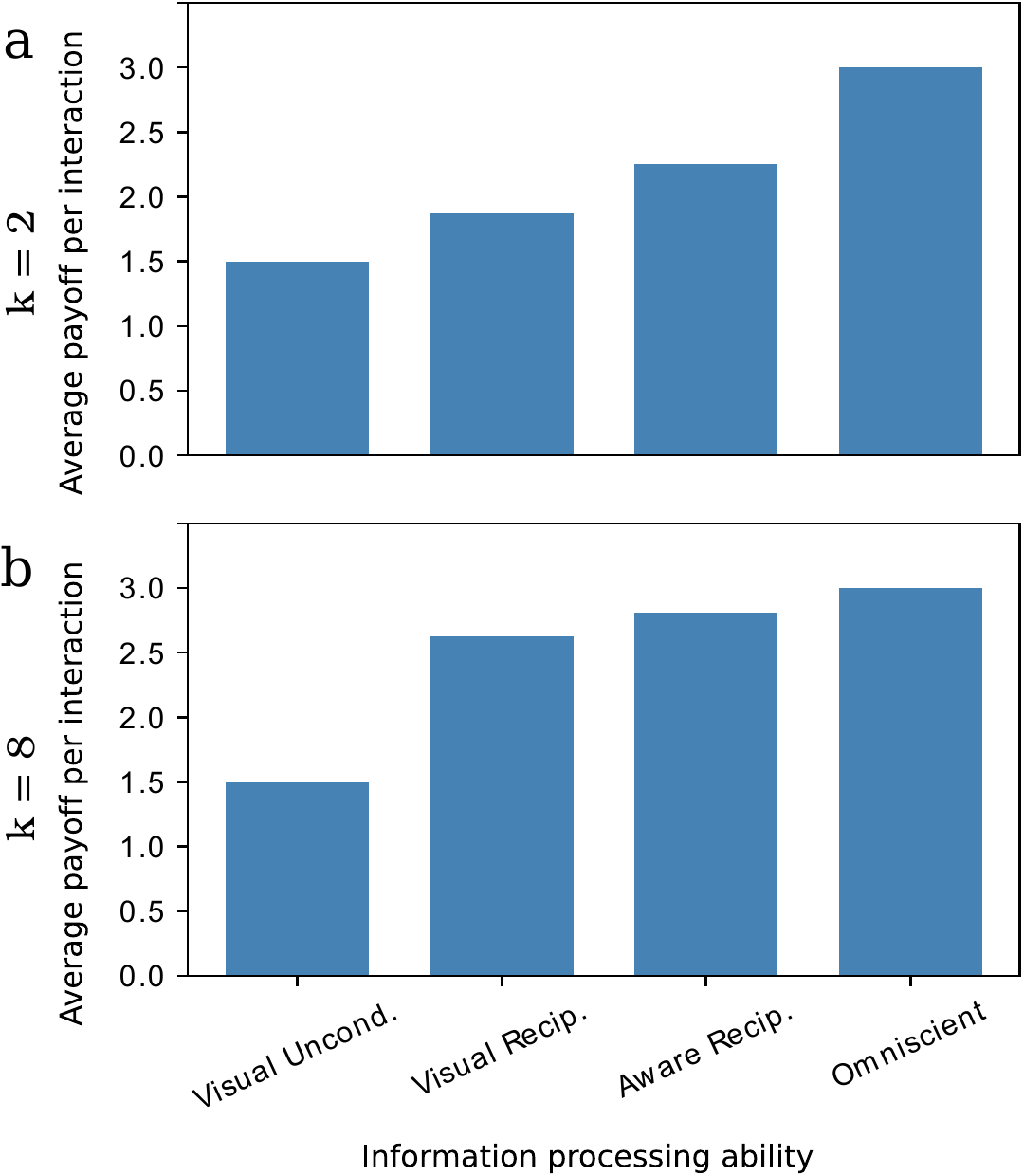}
  \caption{The relationship between individual behavior and payoff in the analytical model for few iterations ($k = 2$) (a) and more iterations ($k = 8$) (b). The $x$-axis also maps to the ability of the individual to use available information, with the lowest ability being on the left, and the highest on the right. Payoffs are computed with $\rho=0.5, R = 3, T = P = 1, S = 0$. More sophisticated behaviors (and thus, those requiring more sophisticated information processing) achieve higher reward. \label{fig:analytical}}
\end{figure}

The total payoff over $k$ iterations of these strategies are:

\begin{align}
\mathcal{U}_{k}(\pi_{\text{VU}}) & = k \left(\rho R+(1-\rho)S \right) \label{eq:pi_VU} \\
\mathcal{U}_{k}(\pi_{\text{VR}}) & = kR-\frac{1-\rho}{\rho}\left(1-(1-\rho)^{k}\right)\left(R-S\right) \label{eq:pi_VR} \\
\mathcal{U}_{k}(\pi_{\text{AR}}) & = (k-1)R + \rho R + (1-\rho)S \label{eq:pi_AR} \\
\mathcal{U}_{k}(\pi_{\text{O}}) & = kR \label{eq:pi_O},
\end{align}
which, for sufficiently large $k$, imply
\begin{align}
\mathcal{U}_{k}(\pi_{\text{VU}}) < \mathcal{U}_{k}(\pi_{\text{VR}}) < \mathcal{U}_{k}(\pi_{\text{AR}}) < \mathcal{U}_{k}(\pi_{\text{O}}) \label{eq:four_strategies}
\end{align}

Figure~\ref{fig:analytical} shows averages of these equations when the interaction is driven by the payoff matrix (\ref{eq:staghunt}), with the chance of sampling a cooperator $\rho = 0.5$, for $k = 2$ iterations and $k = 8$. For full derivations, please refer to the Supplementary Material.

Our model shows, via the inequalities in~(\ref{eq:four_strategies}), that an individual's payoff increases with its ability to process information about others. Less discriminatory behavior results in higher payoff. Finally, from equations (\ref{eq:pi_VU}--\ref{eq:pi_O}), the higher the number of iterations, $k$, the larger the potential benefit of making partner choices based on actual outcome data, rather than perceptible features (contrast Figure~\ref{fig:analytical}a with Figure~\ref{fig:analytical}b).

\section{Reinforcement learning model \label{sec:rl}}

To test the insights from the theoretical model presented above, we create a multiagent reinforcement learning model that captures the properties of iterated interactions with partner choice. The environment is spatially and temporally extended, and information about the outcomes of interactions with partners and between third parties is perceptual.

\subsection{Boat race environment}

\begin{figure*}
  \centering
  \includegraphics[scale=0.70]{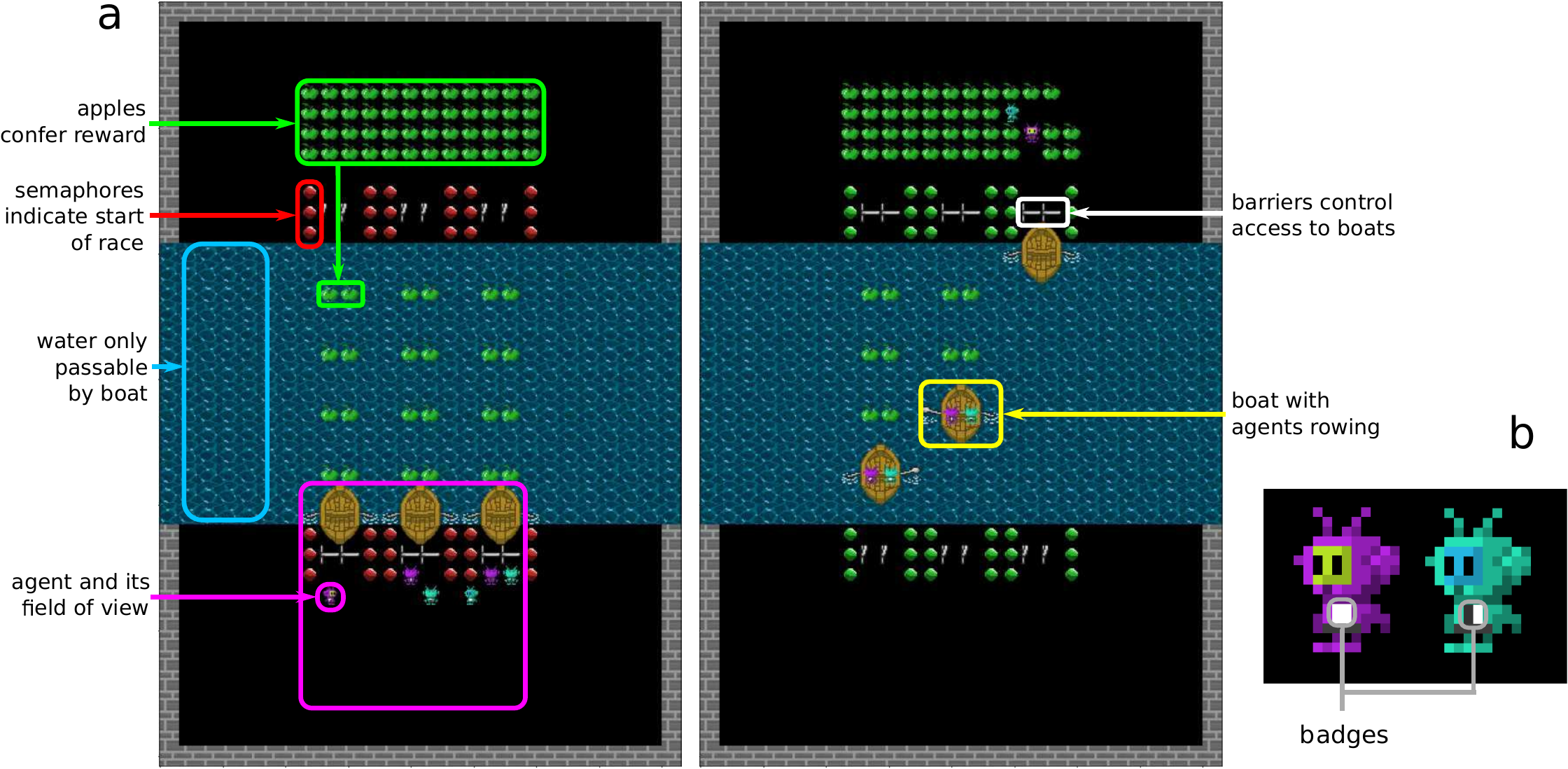}
  \caption{The \emph{boat race} environment. (a) Six players (teal and purple) pair up to row boats (brown) and cross the river to access apples (green) that confer rewards. Agents can move freely on the river banks (black) but cannot walk across the river (water sprites in blue). Access to the boats is gated by barriers (gray). On the left we see a frame from before a race starts (semaphores are red) and agents are able to move freely. On the right we see a frame from after the race has started. The semaphores have turned green, and the barriers lifted to provide access to the boats. (b) Avatars are assigned a unique random ``badge'' at the beginning of an episode so their identity can be known throughout the episode. \label{fig:env}}
\end{figure*}

The environment is inspired by the boat rowing thought experiment posed by Hume~\citep{hume1739treatise}, considered as a golden example of \emph{Stag Hunt}~(See Figure~\ref{fig:env}a). In our environment, 6 players engage in a back and forth series of boat races across a river to reach apples that confer a reward. Boats, however, cannot be rowed by a single player, and thus, players need to find a partner before each race and coordinate their rowing during the race to cross the river. When the players are on the boat, they can choose from two different rowing actions at each timestamp: \emph{paddle}, that is efficient, but costly if not coordinated with its partner; and \emph{flail}, that is inefficient, but isn't affected by the partner's action. When both players \emph{paddle}, the boat moves one cell every $3$ timesteps. When either player \emph{flails}, the boat has a $10\%$ probability of moving one cell, and a reward penalty of $-0.5$ is given to its partner if that partner is currently \emph{paddling} (i.e. if they have executed the paddle action within the last 3 timesteps).

Two perceptible attributes are given to distinguish players~(see Figure \ref{fig:env}b). First, every player is colored purple or teal. The color is readily observable as the players perceive their environment as an RGB image window around themselves (see Figure~\ref{fig:env}a), but is insufficient to individualize a player's behavior. Second, a \emph{badge}, is given to every player at the start of the episode, which uniquely identifies them within the episode. The badge is a pattern of $2 \times 2$ dark gray or white pixels at the center of the player's sprite. The badge is not persistent across episodes.

An episode consists of a number $k$ of discrete races of equal duration. At the beginning of each race, access to the boats is blocked by barriers. The barriers lift after a predetermined amount of time, giving the players time to move freely and approach the boat or partner of their choice (like the ``free mixing'' phase, see Figure~\ref{fig:schematic}). The topology of the environment is such that once a pair of players is behind the barriers of a boat, no other player can force themselves into that boat (this corresponds to the ``partner choice'' phase). If a player so wishes, however, they can back away and walk to another boat. Once a player reaches a seat, they remain seated until either they reach the other river bank, or the time allotted for the race is up, in which case they are disqualified and removed for the rest of the episode (the ``interaction'' phase).

In the environment there are apples that provide reward to the players that touch them. The apples are consumed in the process, but will be replenished eventually (see below). They are positioned in two different locations: along the river to encourage progress; and on the opposite river bank from the players, as a reward for finishing the race (the ``outcome'' phase).

The apples on the river are replenished at the start of each race. The apples on the river bank replenish at a constant rate, but disappear altogether when the race time is up (and re-appear on the opposite bank). The replenishing rate of the apples on the river bank is high enough that not even the $6$ players optimally consuming them can exhaust them. Therefore, there is a strong incentive to finish a race quickly, so as to accrue the largest reward. For full details see the Supplementary Material.

\paragraph{Possible policies.} The environment admits behaviors that implement the policies from the theoretical model. A \emph{visual unconditional} would pair up with agents of one color, when available, or at random when not, and always paddle. A \emph{visual reciprocator} would pair on the first race with an agent of one color if available. If their partner (mostly) paddled during the race, they would remember their badge and pair with them in subsequent races. If the partner (mostly) flailed, they would look for another partner of that color, if available. An \emph{aware reciprocator} would pair on the first race with an agent of one color, if available. During the race, this agent would look at the badges of all the players, and infer whether they are flailing or paddling based on their rowing pattern and boat speed. From the second race, it would exclusively pair with cooperators, if available.

\subsection{Environment validation}

\begin{figure}
  \centering
  \includegraphics[scale=0.5]{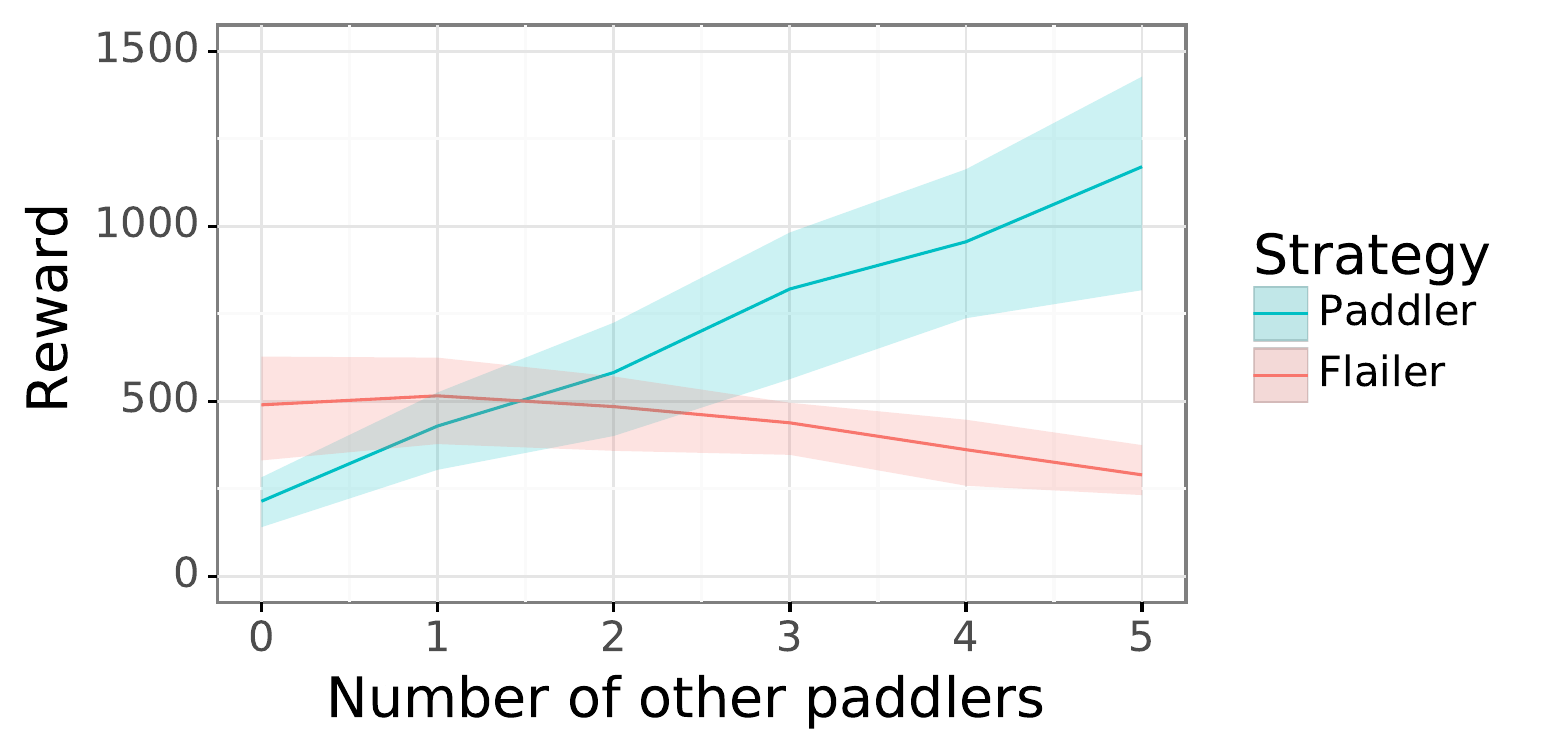}
  \caption{The Schelling diagram of the \emph{boat race} environment with 8 races ($k = 8$), unconditional cooperation (paddler), and defection (flailer). Depicted are the expected payoffs (average in solid line, quartiles in shaded area) that an individual joining an episode with a particular number of paddlers (on the $x$-axis) would obtain, depending on their choice of strategy (paddler or flailer). \label{fig:schelling}}
\end{figure}

To verify that our environment has the properties of an iterated \emph{Stag Hunt}, we compute in Figure~\ref{fig:schelling} its \emph{Schelling diagram}~\citep{schelling1973hockey, perolat2017common}. It is possible to read off various game theoretic properties from a Schelling diagram including whether a game is a social dilemma. A Schelling diagram is a plot that summarizes the incentives of a set of players who face a binary choice of a strategy to pursue. In our case the choice of strategy is either always paddle (paddler), or always flail (flailer). The way to interpret the Schelling diagram is to consider a focal individual joining an episode where there are a certain number of paddlers, and the rest are flailers (for a total of $5$ other players). This focal individual then faces a binary choice of whether to join as a paddler, and receive the payoff that paddlers receive given the group composition, or receive the payoff of flailers. For instance, if the episode has $3$ paddlers, and $2$ flailers already, the focal player joining as a paddler would receive the payoff of the paddler's line at $x=3$ in Figure~\ref{fig:schelling} (on average $\approx 800$ reward), and if joining as a flailer would receive the payoff of the flailer's line at $x=3$ (on average $\approx 490$). Therefore, if there are $3$ paddlers already in the environment, there is an incentive to join as a paddler.

To obtain paddlers and flailers, we train A3C agents~\citep{mnih2016asynchronous, espeholt2018impala} with a preferred rowing type. We give a pseudo-reward of $+5$ for executing their preferred rowing type, and of $-5$ for executing the wrong rowing type. All other rewards are left intact. We validated that paddlers overwhelmingly "paddle" when in the boat, while flailers "flail" (see Supplementary Material).

From Figure~\ref{fig:schelling}, we can see that when paddlers are abundant, paddling is advantageous, however, when paddlers are rare, flailing is advantageous, which correspond to a population version of \emph{Stag Hunt} (as per the definition proposed in~\citep{hughes2018inequity}). Thus, we can refer to paddlers as \emph{cooperators} and flailers as \emph{defectors}.

\subsection{Training and evaluation of agents \label{sec:train_and_eval}}

Now that we know that the boat race environment has the right incentive structure, we proceed to train and evaluate reinforcement learning agents. Like in the theoretical model case, we consider a focal agent training in a \emph{community} of other agents. To form the training community of a focal agent, we freeze the parameters of the agents from the Schelling diagram, and refer to them as \emph{bots}. Bots are of one of four types: purple cooperators, purple defectors, teal cooperators, or teal defectors. We will refer to the focal agent as a \emph{naive learner} to differentiate them from the bots in the community. Naive learners can be of either color, and have no intrinsic preference for either type of rowing.

\paragraph{Training.} A training community consists of $20$ bots: $n$ purple cooperators, $n$ teal defectors, $10-n$ purple defectors and $10-n$ teal cooperators, for $n=0,\ldots,10$. Therefore, there are $11$ possible community compositions. Notice that a training community will always have $10$ cooperators (and defectors), and $10$ purple (teal) bots. What changes between compositions is the statistical association of color with strategy, and we will refer to this as the \emph{community bias}. Naive learners will be assigned to a particular community, and train in episodes with $5$ bots sampled (without replacement) from their training community.

We train naive learners with four different architectures: A3C with and without an LSTM~\citep{espeholt2018impala}, and V-MPO with and without an LSTM~\citep{song2019vmpo}. We chose these architectures to reflect different information processing abilities: LSTM provides agents with memory which can be used to reciprocate previous partner cooperation; V-MPO is a more advanced agent than A3C and is expected to learn better. For each agent architecture we train one naive learner of each color for each of the $11$ possible community compositions. We also have two values for the number of races $k$: $2$ and $8$. Therefore, in total, we train $44$ ($=2\times11\times2$) agents of each architecture. These videos (\textcolor{blue}{\href{http://youtu.be/vweO6k6cx5E}{youtu.be/vweO6k6cx5E}} and  \textcolor{blue}{\href{http://youtu.be/nYbiQyT5Rxs}{youtu.be/nYbiQyT5Rxs}}) show episodes of a V-MPO naive learner with an LSTM for $2$ and $8$ races respectively. To improve interpretability, we post-process the video to highlight the naive learner in white, cooperator bots in blue, and defector bots in red.

\paragraph{Evaluation.} For each naive learner, we ran $50$ \emph{test} episodes with a new \emph{test} community that had no community bias (i.e. color and strategy are uncorrelated). This testing community was composed of held out bots that had trained in the same number of races as the naive learner but it has not interacted with at training. Therefore, any partner choice observed is due to zero-shot transfer to co-players.

The \emph{association matrix} $P$ of a naive learner is a $2 \times 2$ matrix of counts where $P_{i,j}$ corresponds to the number of times the naive learner shared a boat with a bot with color $i\in{p,t}$ (for purple and teal, respectively) and strategy $j\in{c,d}$ (for cooperator and defector, respectively). For instance, $P_{t,d}$ corresponds to the number of times the naive learner associated with a teal defector. We define the \emph{participation} of an agent as the sum of the entries in their association matrix $P$. We define the \emph{discrimination index} of a naive learner as
\[
\mathcal{D} = \left| P_{p,c} - P_{t,c} \right| + \left| P_{p,d} - P_{t,d} \right| - 
              \left| P_{p,c} - P_{p,d} \right| - \left| P_{t,c} - P_{t,d} \right|
\]

This index measures the difference between how much an agent is associating by color: $\left| P_{p,c} - P_{t,c} \right| + \left| P_{p,d} - P_{t,d} \right|$ (sum of absolute differences across rows); versus how much it is associating by behavior: $\left| P_{p,c} - P_{p,d} \right| + \left| P_{t,c} - P_{t,d} \right|$ (sum of absolute differences across columns). A positive value means the agent prefers to associate with bots based on their color rather than their behavior. A negative value means they prefer to associate by behavior rather than color. A value of zero means that they oversample a color, only insofar as they oversample a behavior (i.e. there isn't a row or a column that is larger element-wise than the other, see Supplementary Material).

\paragraph{Estimated CO${}_2$ emissions. } \label{para:co2cost}

Experiments were conducted using an internal GPU cluster with carbon efficiency of 0.27 kgCO$_2$eq/kWh. A cumulative of 30,000 hours of computation was performed on Tesla P100 GPUs (TDP of 250W). Total emissions are estimated at 2025 kgCO$_2$eq, all of which were directly offset. Estimations were conducted using the \textcolor{blue}{\href{https://mlco2.github.io/impact}{Machine Learning Impact calculator}}~\citep{lacoste2019quantifying}.

\section{Results}

To understand the policy that the naive learners are executing, we keep track of their association matrix, and what type of rowing they engage in. Naive learners overwhelmingly paddle, but when paired with a defector, they flail or do not row at all (see Supplementary Material). We also tested if the agents (both naive learners and bots) had any preference for a particular seat (left or right), or boat (left, middle, or right), but they did not (see Supplementary Material). Thus we can infer that any association between particular individuals is a result of explicit partner choice, either of themselves or others.

To understand the extent to which an agent is using information about their partners during an episode, we track the discrimination index on the first and last races. $95\%$ confidence intervals for the discrimination index range from $(-48.0, -28.0)$ for an agent with perfect information who exclusively prefers one strategy, to $(28.0, 48.0)$ for an agent who exclusively prefers one color. A naive learner sampling at random has a discrimination index within $(-10, 10)$ with $95\%$ confidence (see Supplementary Material).

\begin{figure*}
\begin{minipage}[t]{.45\linewidth}
\vspace{0pt}
\centering
\includegraphics[width=1.8in]{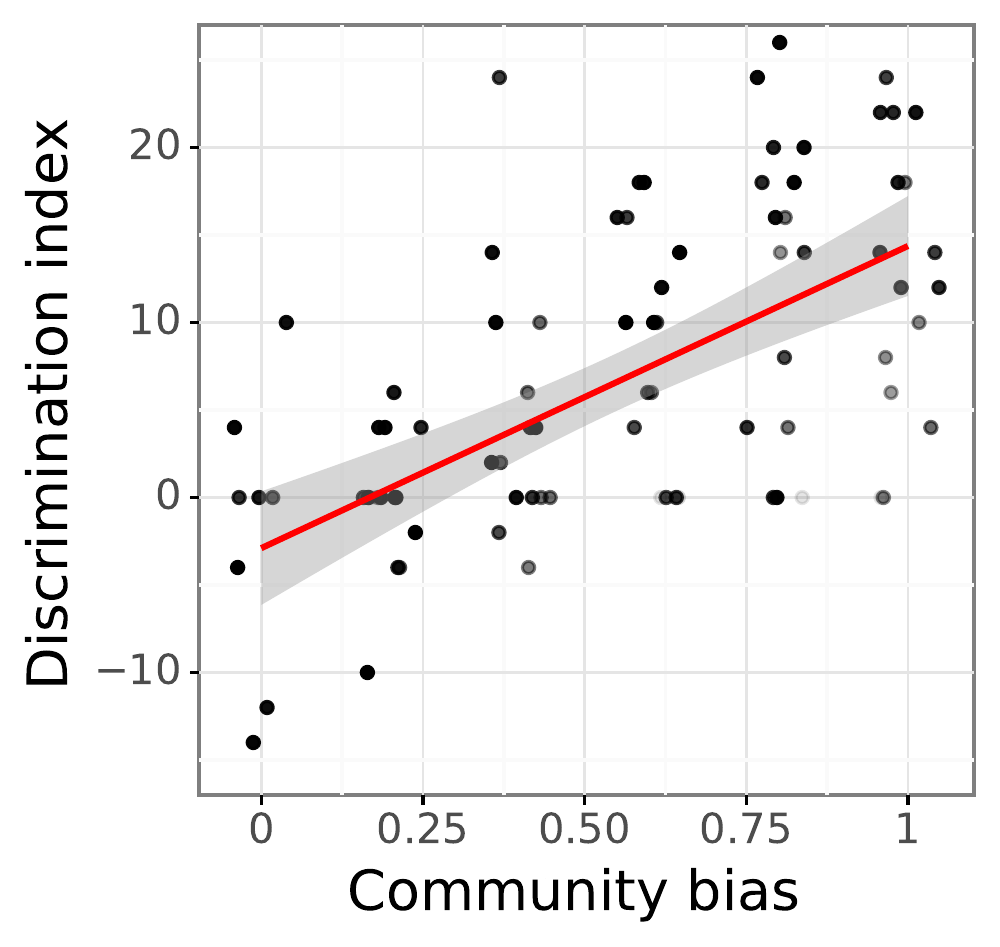}
\end{minipage}%
\begin{minipage}[t]{.5\linewidth}
\vspace{0pt}
\centering
  \begin{tabular}[h]{cccc}
    \toprule
    LSTM &	$k$ &	Race &	Discr. index \\
    \midrule
    No &	2 &	first &	17.838710 \\
    No &	2 &	last &	21.379032 \\
    No &	8 &	first &	8.717742 \\
    No &	8 &	last &	13.153226 \\
    Yes &	2 &	first &	17.750000 \\
    Yes &	2 &	last &	14.129032 \\
    Yes &	8 &	first &	6.798387 \\
    Yes &	8 &	last &	4.629032 \\
    \bottomrule
\end{tabular}
\end{minipage}
  \caption{\textbf{Left:} Discrimination index as a response to training community bias. Plot corresponds to A3C naive learners without an LSTM on their last race of $8$. Other cases show a similar response (see Supplementary Material). Each point shows average values over $50$ episodes in unbiased test communities. The shading of the points represents the naive learner's participation (light is less participation, dark is more). We plot the best fit linear regression (red line) with bootstrapped $95\%$ confidence intervals (gray shaded region). \textbf{Right:} Discrimination index in the first and last races for agents with and without LSTM, for $2$ or $8$ races. Agents trained in $8$ races discriminate less overall, agents without LSTM discriminate more in the last race, while agents with LSTM discriminate less. \label{fig:discr_summary}}
\end{figure*}

To evaluate the effect that the training community bias has on the learned behavior of naive learners, we plot the test time discrimination index, as described in Section \ref{sec:train_and_eval}, as a response of the training community bias that the naive learner experienced. The left panel of Figure~\ref{fig:discr_summary} shows this plot where each point represents averages of $50$ test episodes for a single naive learner. Agents trained in an unbiased community exhibit a slight negative discrimination index in the last race, whereas agents trained in highly biased communities universally have positive discrimination.

Notice that in the case of $k = 8$ races, in principle agents have enough opportunity to sample each of their $5$ co-players each in one race, leaving enough leeway to partner with cooperators in the last race, if any are available. In the case of $k = 2$ races, sampling a cooperator in the last race requires either sampling a cooperator by luck on the first race, and remembering to partner with them in the last race, or, more generally, observing the behaviours of other agents and inferring their policy based on their performance. The environment allows for this type of deduction, as the information is available even though it is challenging to keep track of.

Agents trained in highly biased communities tend to exhibit high discrimination at testing time. This is unlikely only due to competition for the preferred partners since the effect changed with the training community bias, and agents were never observed to associate by behavior as much as they associate by color (their discrimination index was never very negative, see Figure~\ref{fig:discr_summary}, and the Supplementary Material).

The discrimination index of the RL agents increases roughly linearly with their training community's bias (Figure~\ref{fig:discr_summary}, left). This is in contrast to predictions from the analytical model, where any detectable bias should be maximally exploited by reward maximizers. As a consequence, we would have expected more of a threshold response to increases in training community bias.

Only agents with an LSTM were able to have a lower discrimination index in their last race than their first one (Figure~\ref{fig:discr_summary}, right). However, none of our agents were able to robustly ignore the correlation between color and behaviour at test time, even in the most favorable case: a V-MPO agent with LSTM, in $8$ races. Importantly, this is not because agents are unable to effectively realize their partner choice preferences, since agents from highly biased communities were indeed able to associate by color effectively, and often within the confidence interval of optimal association (by color). Agents trained on $8$ races discriminated less than those trained in $2$ races (Figure~\ref{fig:discr_summary}, right). This suggests that the incentives to discriminate can be reduced through repeated interactions.

\section{Limitations} \label{section:limitations}

By design, we limit the scope of this work to that of statistical discrimination. Our work tackles one particular aspect of discrimination that we can study directly with the tools currently available because it does not require predictions or assumptions about multi-generational dynamics on human societies. In doing so, we are able to make a clear and formal contribution, but also limit the possible impact that our insights can have on the societal problems at large. Statistical discrimination is almost certainly a factor within societal discrimination, but the effect size of this factor is unknown.

The results discussed here do not, nor they attempt to, claim to solve the complex problems of racism, sexism, ableism, etc; nor of representation of specific or implied groups of people. These problems require a multi-faceted approach, and no one aspect is likely to resolve them outright. Finally, while reinforcement learning agents are a model of human behavior and learning, they are far from perfect. They are  better able to capture spatial and temporally complex behaviors than traditional rational economic models and evolutionary agent-based models of behavior, but are still far from the richness of the human brain.

\section{Discussion}

Our theoretical model shows that agents have an incentive to statistically discriminate on the basis of perceptible features when their training environment is biased (i.e. the perceptible features are correlated with the quality of social partners). This effect intensifies if they are unable to obtain, process, or incorporate the information of which other agents are good social partners. Agents that choose partners based on their direct quality not only perform better overall, but also discriminate less. This pattern held true in the multiagent reinforcement learning model.

Past research has shown that unfair biases arising in supervised learning models can be mitigated via precautions with training data and model deployment~\citep{mehrabi2019survey, dixon2018measuring,zhang2018mitigating}. Here, we found that reinforcement learning agents trained in a biased community were unable to avoid statistical discrimination. This was expected for agents without an LSTM, for whom reciprocating strategies are likely inaccessible. More surprisingly, it was also true for agents with an LSTM.

Agents with an LSTM are, in principle, able to remember previous social partners in an episode, and thus we expected them to engage in reciprocation, either \emph{visual reciprocation} (associating by color in the first race, but then filter undesirable partners until finding a suitable partner of the preferred color), or \emph{aware reciprocation} (associating by color in the first race, but paying close attention to all interactions and only associating with cooperators in any subsequent races). To our surprise, all agents, regardless of the presence or absence of an LSTM, consistently failed to pair with cooperators even in the last race, maintaining a high discrimination index. This discrimination was not due to an inability to enact partner choices, since agents trained in extremely biased communities exhibited a discrimination index at evaluation that was near the theoretical maximum.

Agents with an LSTM, while unable to fully de-bias themselves, were nonetheless able to discriminate less in the last race than in the first one. This suggests that the ability to process and incorporate information plays a central role in how individuals learn to choose a partner in a non-discriminatory way. This effect, however, was far from optimal, and all agents, with or without an LSTM, behaved closer to the theoretical prediction of \emph{visual unconditional} players (Equation~\ref{eq:pi_VU}) than even \emph{visual reciprocators} (Equation~\ref{eq:pi_VR}). None of them came even close to exhibiting an \emph{aware} partner sampling (Equation~\ref{eq:pi_AR}), even though that information was available in their observations. Mitigating unfair biases in end-to-end reinforcement learning systems may require new safeguards designed with this in mind.

\paragraph{Interventions on the agent.}

It might be possible to modify RL agents to explicitly ameliorate discrimination. However, such interventions would typically require privileged knowledge of the world (for example, the true causal graph of color, behavior, and its influence on reward). The goal of this work is (by choice) not to attempt to solve discrimination, but to propose a framework to study its emergence in learning agents, particularly when they do not have exhaustive mechanisms to deal with discrimination. It is unrealistic to assume that we would be able to enumerate and (manually or automatically) correct all potential sources of bias. We decided to focus on this goal as we believe that shedding light on the emergence of discrimination, an area that is still very poorly understood, is a necessary first step for devising sensible approaches to alleviating this problem.

\paragraph{Fairness in machine learning.}

The study of undesired biases in machine learning systems is an active area of research, ranging from formalizing desiderata for system evaluation to the development of methods for mitigating undesired biases~\citep{barocas2019fairness,chouldechova2020snapshot,gajane2017formalizing, mehrabi2019survey,mitchell2018fair,verma2018fairness}. These efforts have focused predominantly on static prediction settings where only the immediate impact of a decision is taken into account. However, when repeated interactions are not properly accounted for, systems that seem fair in a static sense might exacerbate harmful biases over time~\citep{liu18delayed,zhang2020how}. Awareness of this issue has motivated a surge of research on fairness in sequential decision making settings, including work on bandits and reinforcement learning~\citep{jabbari2017fairness, joseph2016fairness}, as well as on simulation of fairness interventions and causal modelling~\citep{creager2020causal, damour2020fairness}.

Work on fairness in machine learning has made a distinction between \emph{group-level fairness} and \emph{individual-level fairness}. Group fairness ensures some form of parity (e.g. between positive outcomes, or errors) across different protected groups (split by e.g. gender). Individual fairness, in contrast, is concerned with constraints that bind on the level of an individual, for example by requiring ``similar'' people to receive similar outcomes~\citep{dwork2012fairness}. Our work, framed as statistical discrimination, studies how bias emerges and persists and how this violates fairness at the level of an individual. However, while (with some exceptions e.g.~\citep{zhang2014fariness}) most of the literature is concerned with decisions made by one entity (such as e.g. a bank on an individual), this work considers decentralized interactions among peers~\citep{chouldechova2020snapshot}.

\paragraph{Insights.}

Our work suggests two general ways to reduce discrimination: one is to decrease the bias in the training community; and the second is through more careful evaluation of potential partners.

In our setting, theory predicts that small decreases in community bias will have no effect in the policy learned. Individuals will still have an incentive to sample a color exclusively when available. Interestingly, our agents do not conform to that prediction. They instead respond to small reductions in community bias by learning a proportionally less discriminatory policy. We posit that this is due to the stochasticity of sampling co-players for an episode from a community, which results in a bet-hedging solution on the part of the agent. Essentially, whether an agent uses a signal depends on its reliability, and so as the signal becomes more reliable, it becomes more exploitable. This pattern is compatible with the phenomenon observed in the field of ML fairness where reducing bias in the training data has been observed to have a positive impact in the fairness metrics of classification algorithms~\citep{kamiran2012data}. It is also compatible with the \emph{contact hypothesis} which postulates that stereotype-disconfirming interactions reduce prejudice in humans~\citep{allport1954nature, paluck2019contact}.

Our results may also be understood within a dual process framework, which posits that slow and deliberative processes may override automatic habit-like processes. Dual process models have a long history in psychology and neuroscience~(reviewed in \citep{dayan2009goal}). For example in human vision, feedforward decision-making operates rapidly~\citep{thorpe1996speed, greenwald1995implicit} and may learn dedicated threat-related features~\citep{chekroud2014review}, whereas deliberative processes such as mental rotation unfold sequentially via recurrent processing~\citep{lamme2000distinct}, and are consequently much slower~\citep{shepard1971mental}. These differences parallel the contrast between our purely feedforward agents and ones with LSTM.

A related comparison arises between model-based (MB) and model-free (MF) reinforcement learning~\citep{daw2005uncertainty}. Whereas MB learning can prospectively plan to achieve desired outcomes using knowledge of the likely consequences of actions, MF learning gradually updates cached values of actions retrospectively from experience~\citep{dayan2008decision, akam2015}. When outcome values change, model-based decision-makers can rapidly change their behavior, whereas model-free decision-makers need additional repetitions~\citep{dickinson1985actions}. In the human brain, MF learning (for example, of habits and rituals) is associated with the dorsolateral striatum~\citep{graybiel2008habits} while MB learning is associated with persistent activity in the prefrontal cortex~\citep{daw2005uncertainty, wang2018prefrontal}. By allowing agents to learn relevant features from experience when the task demands it, LSTMs may offer a model for the neural mechanisms that underlie MB learning~\citep{wang2018prefrontal}. In our model, agents with LSTMs benefited from incorporating individual-oriented information for partner choice, whereas feed-forward networks were incapable of this and thus continued to discriminate. These observations suggest that engaging different neurological systems may modulate the impact of statistical discrimination.

Discrimination, in all its forms, is a key societal issue. Only by disentangling the factors contributing to its emergence, establishment, and maintenance can we hope to solve this problem. This work contributes to the understanding of the emergence of bias, and serves as evidence that using reinforcement learning systems in combination with traditional analytical tools can provide insights into important societal issues.

\bibliographystyle{ACM-Reference-Format} 
\bibliography{discrimination}

\section{Supplementary material}

\subsection{Analytical model}

We assume a population of individuals that engage in pair-wise
repeated interactions for up to $k$ steps. Each individual has a
perceptible attribute $c\in\mathcal{C}$, a partner sampling
policy $\sigma$, and a behavior policy $\pi$. We assume the set of perceptible
attributes $\mathcal{C}$ is discrete, and will be referred to as \emph{colors}.
Both the sampling policy $\sigma$ and the behavior policy $\pi$ can depend
on the perceptible attributes, the history of past
interactions with a particular individual, or even on hidden information like the behavior policy of other individuals.

In each interaction, individuals play a symmetric game with
two strategies $C$ and $D$ corresponding to \emph{cooperation} and
\emph{defection}, respectively. The payoff of the game is given by
a matrix with two rows ($C$ and $D$) corresponding to the strategies
available to the focal agent, and two columns ($C$ and $D$ also)
corresponding to the strategies of the co-interactant. The payoff matrix
is given by 
\[
\begin{array}{cc}
 & \begin{array}{cc}
C & D\end{array}\\
\begin{array}{c}
C\\
D
\end{array} & \left(\begin{array}{cc}
R & S\\
T & P
\end{array}\right)
\end{array}
\]
where $R$ represents the \emph{reward} of mutual cooperation, $S$
represents the \emph{sucker}'s reward, $T$ represents the \emph{temptation
}to defect, and $P$ is the \emph{punishment} reward for mutual defection~\citep{rapoport1965prisoner}.
In social dilemmas, there is an incentive to defect, either as a temptation
from mutual cooperation, or as risk-avoidance from mutual defection,
or both. For Stag Hunt, $R>T\ge P>S$, which implies there is no temptation
to defect once there is mutual cooperation, but there is a risk-avoidance
incentive to defect. This is sometimes referred to as a lack of \emph{greed}
and presence of \emph{fear}~\citep{macy2002learning}. Notice that the payoff matrix need not be
symmetric. The term symmetric in \emph{symmetric game} refers to the symmetry
of player roles and strategy payoff, rather than the properties of the matrix
itself.

\subsubsection{Behavior policy}

The behavior policy $\pi$ determines both if the relationship should
continue or end; and the strategy to play in the next interaction. When the relationship
ends, a new partner will be sampled and this interaction will not
count towards the $k$ steps. Formally, 
\[
\pi:\mathcal{H} \rightarrow \Delta \{C,D,\Omega\}
\]
where $\mathcal{H}$ is the set of all possible information states available to the individual, and (as discussed above) can include histories of interactions with current and past partners, revealed information about non-partners, and even hidden information about individuals. And $\Delta\{C,D,\Omega\}$ is the set of probabilities over
the two strategies $C$ and $D$, and the termination action $\Omega$. We use the notation $\pi(h)_{X}$
to denote the probability of playing strategy $X$ given the history
$h$ and the color of the partner $c$. In practice, we often consider
deterministic policies, in which case we use the strategy itself as
the output of the policy (e.g. $\pi(h)=C$). The special action
$\Omega$ denotes the end of a relationship.

\subsubsection{Partner sampling policy}

A partner sampling policy $\sigma$ determines the next individual to interact with. This function can represent a completely uniform sample over all individuals, or sampling exclusively individuals with a particular color, and even an oracle that always samples a cooperator. We will restrict what information can be used in a sampling function as a stand-in for an individual's ability to process and incorporate information. So, an individual that has no memory will only have access to the color of other individuals, while an omniscient individual will have access to even the private information of other individuals from the start. There are many other possibilities in between, and we will discuss some later on.

In many cases, we will assume that every time an individual samples a new interaction
partner, it is one they have never interacted with before\footnote{This simplification enables us to avoid keeping track of the histories of previous interactions inducing a more tractable model. This simplification corresponds to an extreme situation in which when once somebody you have interacted with proves themselves socially-unreliable, you don't ever want to interact with them again.}. Formally,
$\sigma$ is a function mapping populations to distributions over populations, i.e. 
\[
\sigma:\mathcal{H} \rightarrow \Delta \mathcal{P}
\]
where $\Delta X\coloneqq\{(z_{x})_{x\in X}:\sum_{x\in X}z_{x}=1,z_{x}\in[0,1]\forall x\in X\}$
is the set of all probability distributions over the discrete set $X$. Thus, we use $\sigma(\cdot)_p$ to denote the probability of sampling individual $p$.

We define $\rho$ as the probability that a sampled individual will cooperate in the interaction. Formally, for a focal individual with information state $h$ let $\mathcal{G}(h) = \{ p \in \mathcal{P} | \pi_p (h') = C \}$ be the subset of the population that will cooperate in the next interaction with the focal individual. Then
\[
\rho = \sum _ {p \in \mathcal{G}(h)} \sigma(\cdot)_p
\]

Individuals don't have access to $\rho$, but it can be used to abstract out all the complexities of partner choice and express an individual's payoff in its terms.

In practice, we will restrict ourselves to 3 different partner sampling functions:

A \emph{visual} function which takes into account only the color of the individuals in the population. Always samples partners of the color with the highest probability of cooperation (i.e. $\arg \max_c \rho$ for $\mathcal{G}(c)$).

A \emph{aware} function that samples like the visual function on the first interaction, but on any subsequent partner samplings, it selects always a cooperator (i.e. $\rho$ is the same as the one for visual on the first interactions, and $\rho = 1$ for all interactions after the first one). This is representing an individual who is initially unaware of the other individuals' policies, but after one interaction with \emph{any} individual, it observes the policy of \emph{all} individuals. This is an extreme case of an individual that is not cheating (in the sense of having access to hidden information of other individuals), but is able to observe and infer correctly all information of all other interactions in the population at the first opportunity.

Finally, an \emph{omniscient} function that always samples a cooperator (i.e. $\rho = 1$ always). This is a cheating individual, and is used only to provide the upper bound on possible performance.

\subsubsection{Broad behavior classes}

For the purposes of this article, we will focus on assessing only
two extreme classes of behaviors. \emph{Unconditional behaviors},
where the information state comprises only the color $c$ (i.e. $\pi_{U}$ is only a function of $c$). And \emph{reciprocating
behaviors}, where the always starts with cooperation with new relationships, and ends the relationship whenever the partner defects,
that is, the strategy ignores color (i.e. $\pi_{R}$ is only a function of the partner's strategy in the previous interaction).
Formally, 
\begin{align*}
\pi_R(\emptyset) & =C, \\
\pi_R(h) & =\begin{cases}
C & \text{if }h_{-1}\ne D\\
\Omega & \text{otherwise}
\end{cases}\text{; for }h\ne\emptyset,
\end{align*}
where $\pi(\emptyset)$ is the strategy to play in a new interaction (i.e.
one with empty history), and $h_{-1}$ denotes the strategy played by the
partner in the previous interaction.

\subsubsection{Individual payoffs}

We now focus on the payoff of a focal individual and find out under
which conditions each of the broad behaviors above are favoured.
Let $\mathcal{U}_{i}(h,\sigma,\pi;h',\sigma',\pi')$ be the payoff
at step $i$ of an individual with information state $h$, sampling policy $\sigma$,
and behavior $\pi$ with a partner with information state $h'$ and policies
$\sigma'$ and $\pi'$. Often, we will omit some or all of the parameters
to $\mathcal{U}_{i}$ when they are clear from context. We can write
$\mathcal{U}_{i}$ recursively as
\[
\mathcal{U}_{i}=\mathcal{U}_{i-1}+\begin{cases}
\pi'(h')_{C}\pi(h)_{C}R+\pi'(h')_{C}\pi(h)_{D}T+\\
\pi'(h')_{D}\pi(h)_{C}S+\pi'(h')_{D}\pi(h)_{D}P & \text{if continuing with same partner}\\
\\
\rho\left[\pi(h)_{C}R+\pi(h)_{D}T\right]+\\
(1-\rho)\left[\pi(h)_{C}S+\pi(h)_{D}P\right] & \text{if resampling partner at step }i
\end{cases}
\]

\paragraph{Unconditional behaviors.}

For unconditional behaviors against partner color $c$ we have two
possibilities, either unconditional defection $\pi_{UD}=D$ or
unconditional cooperation $\pi_{UC}=D$.

If the partner is also using an unconditional behavior using visual partner sampling, the payoffs simplify
to
\begin{align}
\mathcal{U}_{k}(\pi_{UD}) & =\rho kT+(1-\rho)kP \label{eq:pi_UD} \\
\mathcal{U}_{k}(\pi_{UC}) & =\rho kR+(1-\rho)kS \label{eq:pi_UC}.
\end{align}

Observe that whether cooperation or defection yields higher payoff
depends solely on $\rho$ (and thus, indirectly on the sampling function
$\sigma$ and the color of the partner) and the payoff matrix. An individual would prefer to unconditionally cooperate whenever 
\[
\mathcal{U}_{k}(\pi_{UC})>\mathcal{U}_{k}(\pi_{UD})
\]
which, by combining Equations (\ref{eq:pi_UD}) and (\ref{eq:pi_UC}), for
unconditional behavior partners happens if and only if
\[
\frac{\rho}{1-\rho}>\frac{P-S}{R-T}
\]

This inequality has some interesting properties. The left hand side
corresponds to the odds ratio of encountering a cooperator versus a defector
when sampling partners. The right hand side is the ratio of payoffs
associated with cooperation and defection, against an unknown strategy.
This payoffs ratio is the threshold beyond which cooperation is favoured.
We refer to this quantity $\frac{P-S}{R-T}$ as the \emph{stakes of
interaction}. For Stag Hunt, where $R>T$ and $P>S$, the stakes are
always positive. Since $0\le\rho\le1$, there is always some value
of $\rho$ such that the inequality is satisfied. Note that for the
Prisoner's Dilemma, the stakes are negative, which flips the direction
of the inequality and makes it unsatisfiable.

Even within the family of unconditional behaviors, it would be payoff-maximising
to play different strategies against different colored partners,
depending on whether the odds ratio of finding a cooperation ($\rho/(1-\rho)$)
of that color exceed the stakes of the interaction. We will denote
this unconditional strategy that picks the best of $C$ or $D$ for
each color as $\pi_{U}$. Additionally, the payoff-maximising partner
sampling function is simply to sample from the color that has the
highest empirical proportion of initial cooperators $\sigma_{U}=\arg\max_{c}p_{c}$.

In practice, we will just assume an unconditional cooperator, or unconditional defector policy.

\paragraph{Reciprocating behaviors.}

From here on we assume that partnerships are never ended when agents
are in mutual cooperation. This is a reasonable assumption when the
dynamics of interactions are governed by a Stag Hunt payoff matrix,
because mutual cooperation is both a Nash equilibrium and the collective
payoff maximising outcome. Therefore, there isn't any incentive from
any partner to end the relationship. And since reciprocators end the
relationship immediately upon defection of their partner, this is
equivalent to having the choice of partnership ending only on the focal
individual, which simplifies the derivations substantially.

If an individual is using a visual sampling function, we can expand and simplify the recursive equation into 
\begin{align*}
\mathcal{U}_{k} & =\rho kR+(1-\rho)\left[S+\rho(k-1)R+(1-\rho)\left[S+\rho(k-2)R+\ldots\right]\right]\\
 & =\rho R\left(\sum_{i=0}^{k-1}(k-i)(1-\rho)^{i}\right)+(1-\rho)S\left(\sum_{i=0}^{k-1}(1-\rho)^{i}\right)\\
 & =\rho R\left(\sum_{i=0}^{k-1}(k-i)(1-\rho)^{i}\right)+(1-\rho)S\left(\frac{1-(1-\rho)^{k}}{\rho}\right)
\end{align*}

We can simplify the coefficient of the $\rho R$ term as follows 
\begin{align*}
\sum_{i=0}^{k-1}(k-i)(1-\rho)^{i} & =\sum_{i=0}^{k-1}\sum_{j=0}^{i}(1-\rho)^{j}\\
 & =\sum_{i=0}^{k-1}\frac{1-(1-\rho)^{i+1}}{\rho}\\
 & =\frac{1}{\rho}\left(k-(1-\rho)\frac{1-(1-\rho)^{k}}{\rho}\right)
\end{align*}
which yields the final equation
\begin{align}
\mathcal{U}_{k}=kR-\frac{1-\rho}{\rho}\left(1-(1-\rho)^{k}\right)\left(R-S\right)
\end{align}

If an individual is using an aware sampling function, we simplify further as 
\begin{align}
\mathcal{U}_{k} & = (k-1)R + \rho R + (1-\rho)S
\end{align}

And with aware sampling, we obtain the theoretical maximum payoff of
\begin{align}
\mathcal{U}_{k} & = kR
\end{align}

\subsubsection{Behavioral dominance}

Using the computed payoffs of a focal individual pursuing either
unconditional or reciprocating behaviors, we now determine the conditions
for reciprocating behaviors to dominate unconditional behaviors,
as well as the optimal partner sampling policy for these behaviors.

Notice that the payoff for an omniscient reciprocator is greater than that of an aware reciprocator which in itself is greater than that of a visual reciprocator. Therefore, we only need to compare the payoff of an unconditional behavior with a visual reciprocator.

Formally, reciprocating dominates unconditional behaviors if and
only if
\begin{align*}
\mathcal{\mathcal{U}}_{k}(\sigma,\pi_{R}) & >\mathcal{\mathcal{U}}_{k}(\sigma',\pi_{U})\iff\\
kR-\frac{1-\rho}{\rho}\left(1-(1-\rho)^{k}\right)\left(R-S\right) & >\rho'kT+(1-\rho')kP\text{; and}\\
kR-\frac{1-\rho}{\rho}\left(1-(1-\rho)^{k}\right)\left(R-S\right) & >\rho'kR+(1-\rho')kS
\end{align*}
Now, if there is at least one color with an odds ratio of cooperators
above the stakes, then we need only look at the second inequality,
which simplifies to

\begin{align*}
kR-\frac{1-\rho}{\rho}\left(1-(1-\rho)^{k}\right)\left(R-S\right) & >\rho'kR+(1-\rho')kS\iff\\
k\left(1-\rho'\right)\left(R-S\right) & >\frac{1-\rho}{\rho}\left(1-(1-\rho)^{k}\right)\left(R-S\right)\iff\\
k & >\frac{1-(1-\rho)^{k}}{\rho}\cdot\frac{1-\rho}{1-\rho'}
\end{align*}
which is satisfied for sufficiently large $k$.

\subsection{Reinforcement learning model}

We build an environment called \emph{boat race} using DMLab2D~\citep{beattie2020deepmind}, which is a configurable and performant library for creating multi-agent 2D environments. The environment consists of objects and avatars, where each avatar is controlled by a single agent, and the avatars can be used to interact with objects. We will refer to an agent controlling an avatar as a player, and we will refer to a player's actions or observations as they pertain to their avatar.

An episode on the environment consists of a number of races $k$ that is either $2$ or $8$. Each race is separated in $3$ phases, a partner choice phase lasting $65$ steps, a semaphore changing phase lasting $5$ steps, and a rowing phase lasting $230$ steps. Each race runs in the opposite direction of the previous one, alternating North and South directions.

The objects in the environment are:

\begin{itemize}
    \item \textbf{Apples}. Players entering the same location as the apple consume the apple and receive a reward. The apple will respawn depending on its location: at the end of a race for apples above the river; and with a probability of $10\%$ per step for the apples on the goal river bank.
    \item \textbf{Barriers}. Barriers prevent players from entering their location, and are used to gate access to the boats before the rowing phase of the race starts.
    \item \textbf{Semaphores}. These are impassable objects that show a traffic-light pattern signaling the transition from the partner choice phase to the rowing phase of a race. They are colored red during partner choice, and change to yellow for the $5$ steps of the semaphore changing phase, and then to green in the rowing phase of the race.
    \item \textbf{Boat}. Boats have two seats that players can enter. Once a player enters a seat, they control the oar next to it with their rowing actions. Players in a seat cannot move anymore, and all of their movement actions are no-ops. Boats can only move during the rowing phase of a race, and require two players, one on each seat to move. When a boat reaches the opposite river bank, it automatically disembarks its players into the river bank, past the barriers.
    \item \textbf{Water.} Water is impassable by players on foot (i.e. not in a boat). Boats with players can travel across water.
\end{itemize}

Episodes start with boats on the South river bank, with no apples on that bank, and its barriers closed. The North bank contains apples and its barriers are open. Once the rowing phase starts, the barriers toggle with the North ones being blocked and the South ones being open. At the beginning of the next race, apples on river banks also toggle, with the South ones appearing, and the North ones disappearing.

Players have movement actions and rowing actions. The players can move in any of the cardinal directions (North, East, South or West), and can also turn $90$ degrees left or right. There are two rowing actions: paddle and flail. Rowing actions are considered no-ops when the players are not on a boat seat. Similarly, movement actions are no-ops when players are on a seat. Flailing actions can be executed at every time step, and have a $10\%$ probability (non-cumulative for each player) of moving the boat forward at each time step. Paddling actions have a cool down of $2$ steps, meaning the can only be executed every $3$ steps. Paddling during the cool down period are considered no-ops. Paddling by a partner during the cool down phase will result in the boat moving $1$ cell. Flailing by a partner during the cool down phase results in a reward punishment of $-0.5$ and the paddling attempt is wasted, and progress can only be made by flailing until the cool down ends.

Players have partial observability of $11 \times 11$ cells around them, with $5$ cell to each side of their position, $1$ cell behind, and $9$ in front. They perceive their environment as RGB images of size $176 \times 176 \times 3$ (sprites are $16 \times 16$). Players who do not reach the other side of the river by the end of the race are disqualified and removed from the episode. The agent's observations in this case are fully black (zeros) and all their actions are no-ops.

Agents have a persistent color across episodes, which is instantiated as an avatar of that color in the level. Avatars can be of two colors: purple (base RGB~$ = (145, 30, 180)$) and teal (base RGB~$ = (30, 180, 145)$). In addition, the avatar is assigned a random badge for each episode, used to uniquely identify the players, but only within that episode. The badge is a set of $4$ pixels that can be either black or white. The color of the agent is more salient than the badge.

Apples on the river banks respawned at a rate of $0.1$ per timestep, and confer a reward of $+1$ to the player who eats them.

\subsubsection{Agent details}

All agent architectures had the same size convolutional net with two layers with output channels $16$ and $32$, with a stride of $8$ pixels. The convolutional net was followed by a feedforward net with two layers, both with $64$ output units. Agents with LSTM, have a hidden state of size $128$, using an unroll length of $100$. Agents were trained on mini-batches of data of size $16$. V-MPO had a pop-art layer~\citep{hessel2019multi} for normalizing the value function. A3C minimized a contrastive predictive coding loss \citep{oord2018representation} in the manner of an auxiliary objective \citep{jaderberg2016reinforcement}, which in this case contrasted between nearby time points via LSTM state representations (a standard augmentation in recent work with A3C).

Agents for the training community and the production of the Schelling diagram were trained for $3 \times 10^8$ steps, whereas naive learners were trained for $1.5 \times 10^9$ steps. We used a learning rate of $4 \times 10^{-4}$ for A3C agents and $1 \times 10^{-4}$ for V-MPO agents.

\begin{figure}[ht]
  \centering
  \includegraphics[scale=0.5]{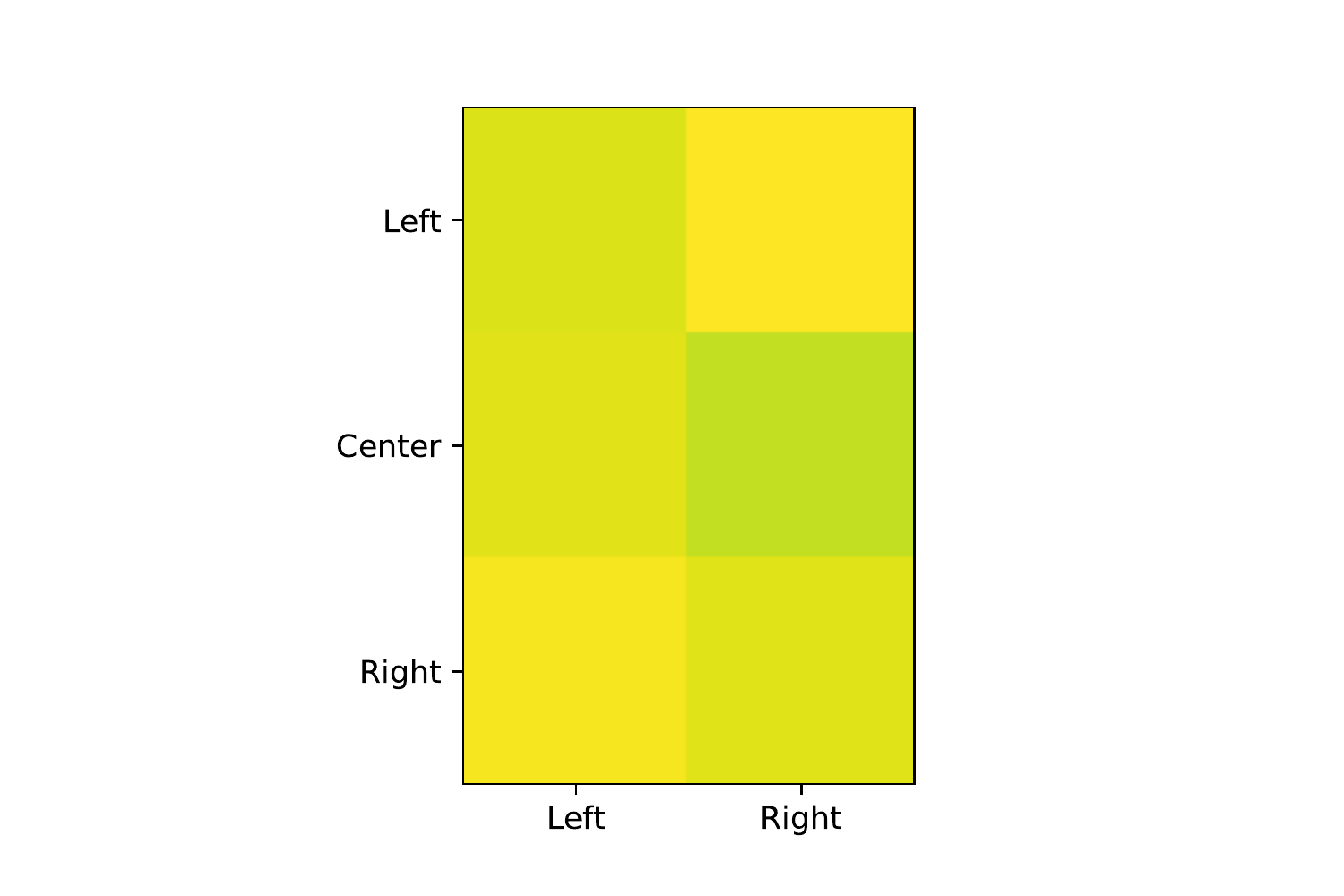}
  \caption{Physical association of community bots who participated in a race. The figure shows a heat map of the counts for each time a bot was on each of the boats (left, center, right) taking one of the seats (left, right). The values ranged from $220$ (darker), to $232$ (lighter). \label{fig:physical}}
\end{figure}

Agents for the training community and Schelling diagram were incentivized to either unconditionally paddle or unconditionally flail. These agents receive a reward of $+5$ for every rowing that matches their incentive, and $-5$ if it didn't. Agents overwhelmingly learned to use their incentivized rowing. Across $350$ episodes, paddlers never flailed in either $2$ races or $8$ races. Flailers paddled only twice out of over $200,000$ flailing actions, and only in the case of $2$ races. Community bots did not exhibit any significant preference towards a particular boat, or a particular side seat within the boat (see Figure \ref{fig:physical}).

Out of $2,200$ episodes used for evaluation for each architecture, we counted the number of times the naive learner paddled and flailed (see Table \ref{tab:paddle_flail}). Given that paddling has a cool down, flailing would be expected to be 3 times as frequent if agents were rowing at random. Naive learners primarily learned to paddle, with the weakest case being the A3C agent with LSTM, where the absolute flailing was greater than the absolute paddling, but only by a factor of two. We also calculated the correlation of the paddling and flailing on the first race of the naive learner with its partner. Overall, there was a positive correlation between the flailing for agents without an LSTM. The rest of the correlations were low (see Table \ref{tab:paddle_flail}).

\begin{table}[ht]
    \centering
\begin{tabular}[h]{ccccc}
    \toprule
    Architecture &	\# flail &	\# paddle & flail corr. & paddle corr. \\
    \midrule
    A3C LSTM    &	34,062   & 17,604   & 0.07  & -0.12 \\
    A3C no LSTM &	19,525   & 31,217   & 0.26  & 0.03 \\
    VMPO LSTM   &	8,767    & 20,370   & 0.06  & 0.10 \\
    VMPO no LSTM &	18,120   & 27,756   & 0.21  & -0.02 \\
    \bottomrule
\end{tabular}
    \caption{Counts of the number of paddle actions and the number of flail actions for naive learners as well as the correlation between their rowing and that of their partner. All counts correspond to the first race for a total of $2,200$ episodes where the partners were uniformly random across strategy and color.}
    \label{tab:paddle_flail}
\end{table}

\subsection{Discrimination index}

We define the \emph{association matrix} $P$ of a naive learner as a $2 \times 2$ matrix of counts where $P_{i,j}$ corresponds to the number of times the naive learner shared a boat with a bot with color $i$ and strategy $j$. We define the \emph{participation} of an agent as the sum of the entries in their association matrix $P$. We define the \emph{discrimination index} of a naive learner as
\[
D = \left| P_{p,c} - P_{t,c} \right| + \left| P_{p,d} - P_{t,d} \right| - 
    \left| P_{p,c} - P_{p,d} \right| - \left| P_{t,c} - P_{t,d} \right|
\]

For simplicity, let's rename the entries of $P$ as $a, b, c,$ and $d$. Without loss of generality, $a > b, c, d$, thus there are 4 cases: $c > d \& b > d$, $c > d \& b < d$, $c < d \& b > d$, or $c < d \& b < d$. Note that $D = 0 \iff a > d > b, c$. Otherwise $D = 2(c-b)$ or $2(c-d)$ or $2(d-b)$ which is always even.

\begin{figure}[ht]
  \centering
  \includegraphics[scale=0.4]{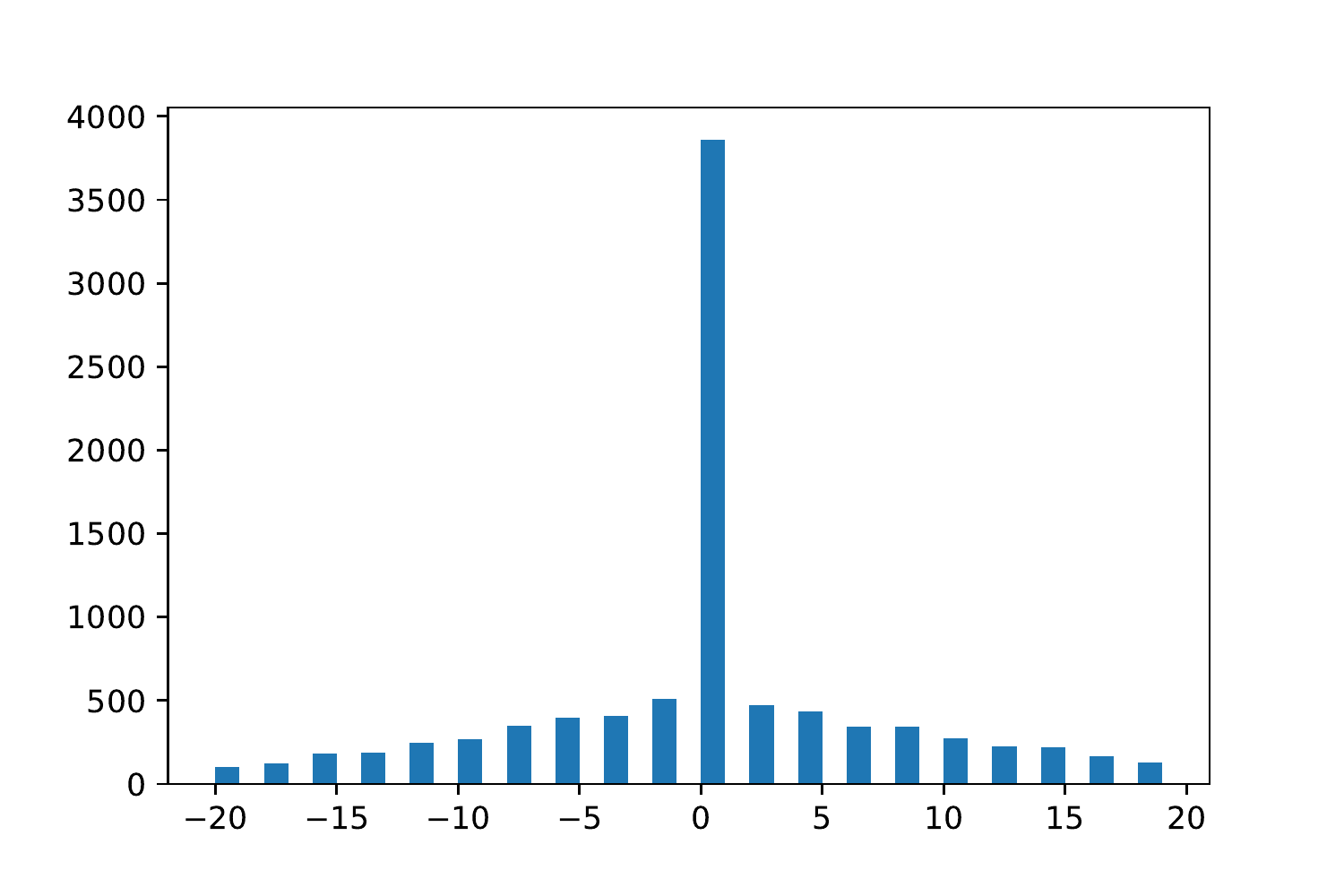}
  \caption{Histogram of discrimination index for $10,000$ simulations of unbiased random sampling of partners. \label{fig:neutral_discr}}
\end{figure}

\begin{figure}[ht]
  \centering
  \includegraphics[scale=0.4]{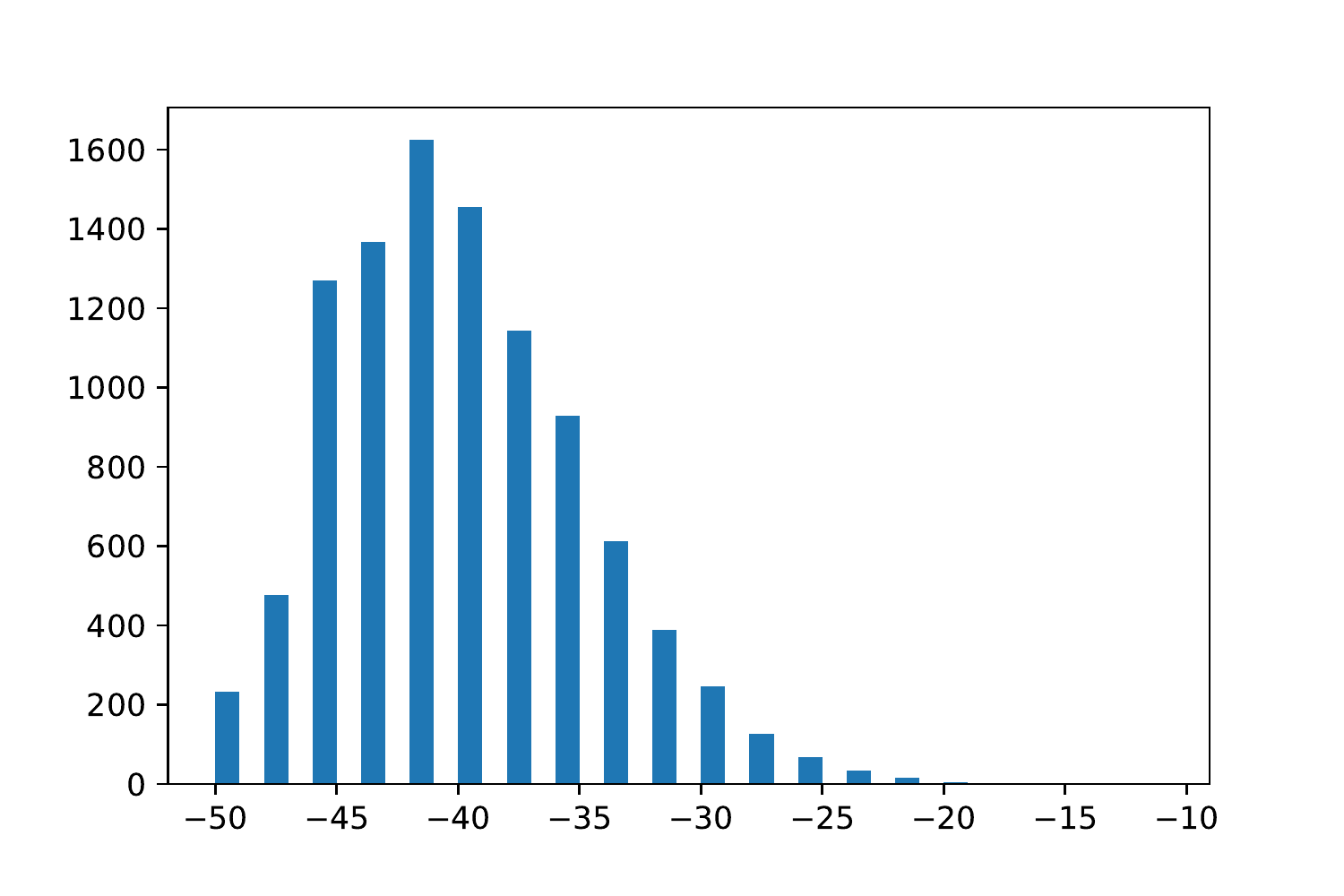}
  \caption{Histogram of discrimination index for $10,000$ simulations of sampling cooperators (uniformly random), if any are available, otherwise, sampling a random partner. \label{fig:perfect_discr}}
\end{figure}

We simulated an individual sampling partners according to a particular strategy. Figures \ref{fig:neutral_discr} and \ref{fig:perfect_discr} show the histogram over $10,000$ simulations of an individual sampling uniformly at random, or sampling only based on one color, when present. We assume that the other $5$ players are of a random color and random strategy, and that the focal individual always has first choice of partner.

\begin{figure}[ht]
  \centering
  \includegraphics[scale=0.5]{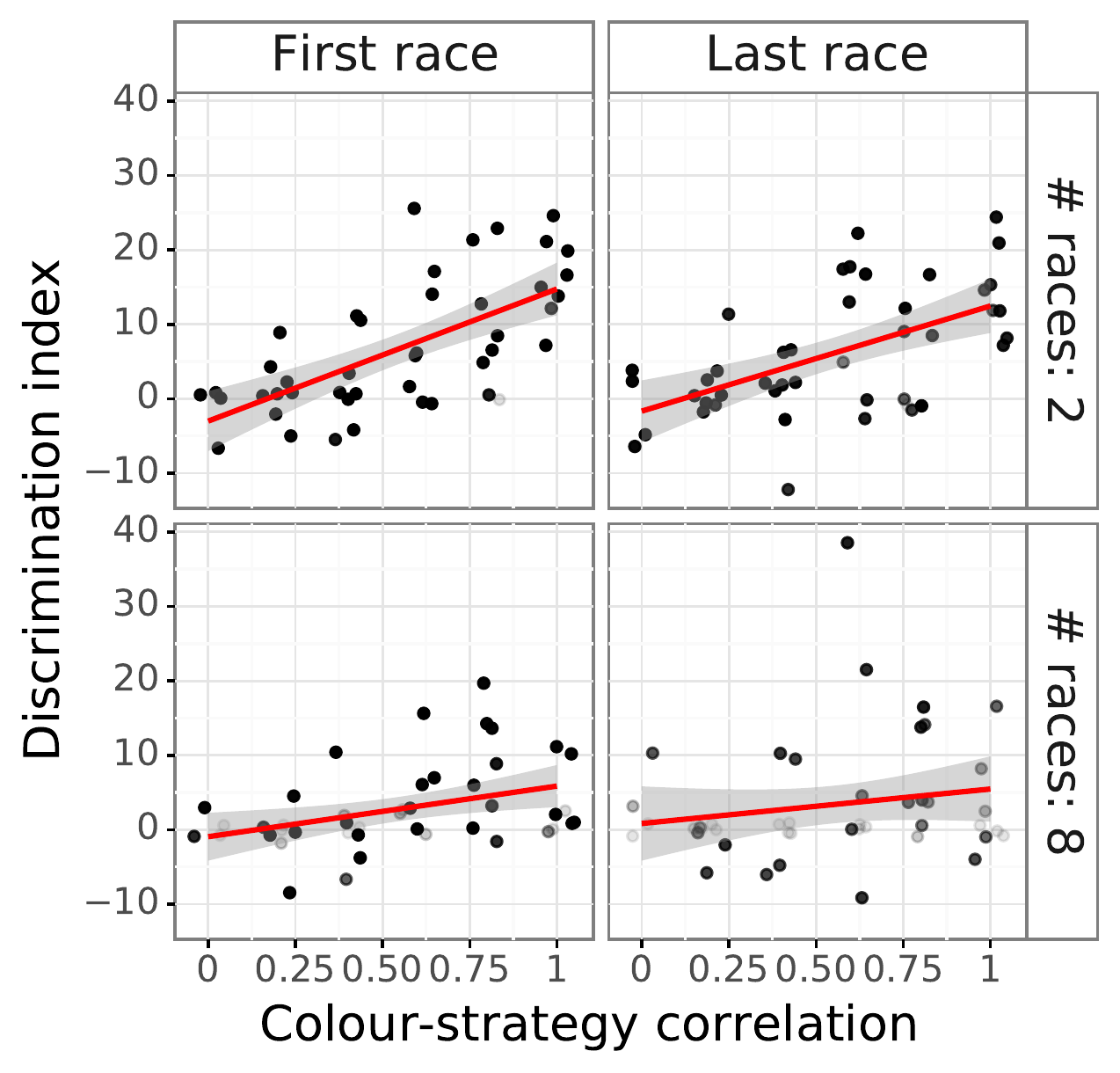}
  \caption{Discrimination index for agents with an LSTM in the first and last races, for the cases of $2$ and $8$ races. \label{fig:discr_plots_lstm}}
\end{figure}

\begin{figure}[ht]
  \centering
  \includegraphics[scale=0.5]{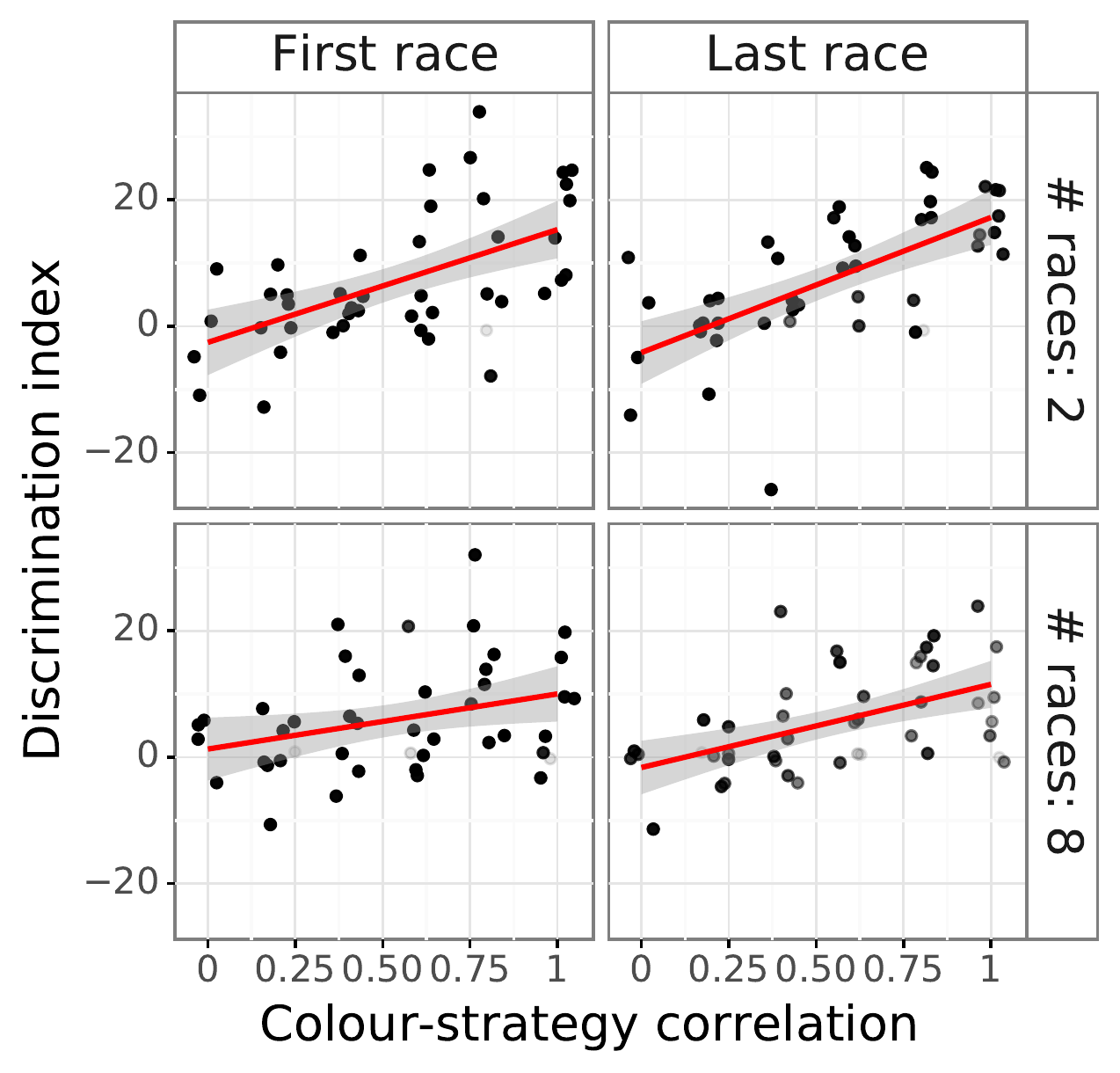}
    \caption{Discrimination index for agents without an LSTM in the first and last races, for the cases of $2$ and $8$ races. \label{fig:discr_plots_no_lstm}}
\end{figure}

Figures \ref{fig:discr_plots_lstm} and \ref{fig:discr_plots_no_lstm} show the discrimination index as a value of the training community bias, as measured in an unbiased evaluation community. A clear pattern of higher community bias resulting in higher discrimination index is seen throughout.

\end{document}